\documentclass[letterpaper]{article} 
\usepackage{aaai2026}  
\usepackage{times}  
\usepackage{helvet}  
\usepackage{courier}  
\usepackage[hyphens]{url}  
\usepackage{graphicx} 
\usepackage{natbib}  
\usepackage{caption} 
\usepackage{tikz} 
\usetikzlibrary{positioning, arrows.meta}
\usepackage{amsmath}
\usepackage{amssymb}
\usepackage{algorithm}
\usepackage{algorithmic}

\usepackage{newfloat}
\usepackage{listings}
\DeclareCaptionStyle{ruled}{labelfont=normalfont,labelsep=colon,strut=off} 
\floatstyle{ruled}
\newfloat{listing}{tb}{lst}{}
\floatname{listing}{Listing}
\title{NMINE: Normalized Mutual Information Neural Estimation}
\author{ Petra Eerikinharju, Marko Tuononen, Ville Hautamäki } \affiliations{ University of Eastern Finland}

\begin{document}

\maketitle

\begin{abstract}
Mutual information is a general measure of statistical dependence that captures both linear and nonlinear relationships between random variables. For continuous and multidimensional variables, however, mutual information is difficult to compute exactly and must be estimated from samples. Moreover, because mutual information is unbounded and expressed in units that depend on the underlying variables, its raw values are not directly comparable across datasets, dimensions, or applications. Normalized mutual information addresses this limitation by converting mutual information into a normalized dependency score. Recent work has demonstrated the practical value of normalized mutual information in applications such as molecular dynamics \cite{nagel2024accurate} and interpretable machine learning \cite{tuononen2025receiver}, but existing estimators remain sensitive to dimensionality and numerical stability \cite{tuononen2025stability}.

In this paper, we propose a fully neural normalized mutual information estimator for continuous variables. The proposed approach combines a MINE-based neural mutual information estimator \cite{belghazi2018mine} with MI-NEE-inspired neural marginal entropy estimators \cite{chan2019nee}. Mutual information is estimated using the Donsker--Varadhan variational representation as implemented in MINE, while marginal entropies are estimated by learning the divergence between each marginal distribution and a uniform reference distribution, from which entropy is recovered. The resulting estimator provides a neural alternative to k-nearest-neighbor-based normalized mutual information estimation \cite{kraskov2004estimating,nagel2024accurate}.

Experiments on Gaussian data from one to eight dimensions show that the proposed estimator improves accuracy over a KSG-based normalized mutual information baseline. These results indicate that neural estimation is a promising direction for normalized dependency measurement in continuous multidimensional settings.
\end{abstract}


\section{Introduction}

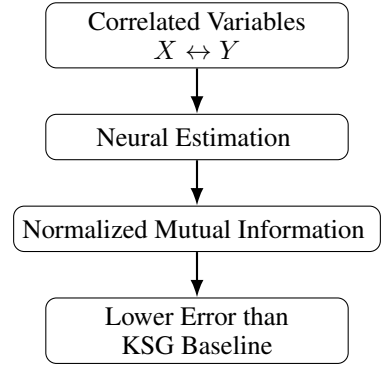
\begin{figure}[t] \centering \begin{tikzpicture}[ scale=0.60, node distance=0.6cm, >=Latex, box/.style={ draw, rounded corners, minimum width=4cm, minimum height=0.6cm, align=center } ] \node[box] (input) { Correlated Variables\\ $X \leftrightarrow Y$ }; \node[box,below=of input] (neural) { Neural Estimation }; \node[box,below=of neural] (nmi) { Normalized Mutual Information }; \node[box,below=of nmi] (result) { Lower Error than\\ KSG Baseline }; \draw[->,thick] (input) -- (neural); \draw[->,thick] (neural) -- (nmi); \draw[->,thick] (nmi) -- (result); \end{tikzpicture} \caption{ Conceptual overview of the proposed NMINE method. Correlated variables are processed using neural mutual information and entropy estimators to produce a normalized mutual information estimate with lower error than a KSG-based baseline. } \label{fig:visual_abstract} \end{figure}

Mutual information is a general measure of statistical dependence that captures both linear and nonlinear relationships between random variables~\cite{shannon1948mathematical,cover2006elements}. For continuous and multidimensional variables, however, mutual information is difficult to compute exactly and must be estimated from samples~\cite{paninski2003estimation,kraskov2004estimating}. Moreover, because mutual information is unbounded and expressed in units that depend on the underlying variables, its raw values are not directly comparable across datasets, dimensions, or applications. Normalized mutual information addresses this limitation by converting mutual information into a normalized dependency score \cite{vinh2010information,nagel2024accurate}. Recent work has demonstrated the practical value of normalized mutual information in applications such as molecular dynamics \cite{nagel2024accurate} and interpretable machine learning \cite{tuononen2025receiver}. Related information-theoretic measures have also been widely applied in representation learning, disentangled representation learning, and multimodal data fusion, where dependency estimation is used to characterize and regularize latent representations \cite{sanchez2020disentangled,oord2018representation,hjelm2019learning}. However, existing estimators remain sensitive to dimensionality and numerical stability \cite{tuononen2025stability}.

Despite recent progress in neural mutual information estimation, comparatively little attention has been paid to normalized mutual information estimation for continuous multidimensional variables. Existing NMI estimators rely predominantly on nearest-neighbor statistics and may suffer from numerical instability and decreased accuracy as dimensionality increases \cite{tuononen2025stability,nagel2024accurate}.

In this work, we propose a fully neural normalized mutual information estimator that combines Mutual Information Neural Estimation (MINE) \cite{belghazi2018mine} with neural entropy estimation inspired by Neural Entropic Estimation (NEE) \cite{chan2019nee}. We hypothesize that replacing nearest-neighbor estimation with neural divergence estimation improves normalized mutual information estimation accuracy in multidimensional continuous settings.

The proposed framework replaces nearest-neighbor estimation with neural divergence estimation in both the mutual information and entropy components.

The main contributions of this paper are:

\begin{itemize}
\item We introduce a fully neural estimator for normalized mutual information between continuous multidimensional variables.
\item We combine MINE-based mutual information estimation with MI-NEE-inspired entropy estimation in a unified framework.
\item We derive entropy estimates from neural divergence estimates relative to uniform reference distributions and use them to construct normalized mutual information estimates.
\item We evaluate the proposed method on multidimensional Gaussian data and compare it against a KSG-based NMI baseline.
\end{itemize}

\section{Background}

Mutual information (MI) and its normalized variant (NMI) are fundamental concepts in information theory and machine learning, widely used to quantify statistical dependence between random variables \cite{cover2006elements, belghazi2018mine, oord2018representation}. Unlike correlation, which captures only linear relationships, MI can detect both linear and nonlinear dependencies \cite{kraskov2004estimating}. Consequently, MI has been employed for unsupervised feature selection, extraction, clustering, and representation learning, and it plays an important role in statistics, signal processing, and machine learning \cite{cover2006elements,vinh2010information, hjelm2019learning,chen2016infogan,oord2018representation}. MI is also central in modern generative models, influencing both the quality of learned representations and the effectiveness of reconstruction-based objectives \cite{hjelm2019learning, chen2016infogan}.

In practical scenarios, exact computation of MI and NMI is feasible only for discrete variables or special continuous distributions. In most real-world applications, MI must be approximated using an estimator. MI estimation is a well-known difficult problem, particularly in continuous and high-dimensional settings \cite{paninski2003estimation,mcallester2018limitations}. Classical estimators such as histogram-, kernel-, or $k$NN-based methods tend to break down in high dimensions or when the underlying distributions are complex or nondifferentiable, making them unsuitable for gradient-based optimization. These methods struggle to capture sparse or highly nonlinear interactions among multiple variables \cite{gao2015efficient}. Neural estimators can scale to high-dimensional data and complex distributions, although their statistical properties and optimization stability remain active topics of research \cite{belghazi2018mine,mcallester2018limitations}.

\subsection{Mutual Information}

A fundamental difficulty with MI as a similarity measure is that its numerical range depends on the entropies of the variables being compared, making it difficult to interpret whether a given value is large or small \cite{vinh2010information,nagel2024accurate}. The range of MI satisfies

\[ \begin{aligned} I(X;Y) &\le \min\{H(X),H(Y)\} \\ &\le \frac{1}{2}\bigl(H(X)+H(Y)\bigr) \\ &\le H(X,Y). \end{aligned} \]

Normalizing MI facilitates interpretation and comparison across different conditions by transforming mutual information into a normalized dependency measure \cite{vinh2010information,nagel2024accurate}.

Several normalization schemes have been proposed in the literature. Most classical schemes were developed for discrete variables and clustering evaluation, where entropies are well-defined and bounded \cite{vinh2010information,strehl2002cluster}. In continuous settings, normalization is more subtle due to the lack of invariance of differential entropy and the unbounded nature of MI \cite{jaynes1968prior,nagel2024accurate}. This motivates the need for principled normalization strategies that remain meaningful for continuous variables.

\subsection{Symmetric Normalization}

A key problem with symmetric NMI is that its normalization factor depends on the entropies of the two variables, not solely on their dependence \cite{jerdee2025}. As a result, normalization can change the ranking of candidate models. Even if

\[
I(X_1;Y) > I(X_2;Y),
\]

it may still hold that

\[
\frac{I(X_1;Y)}
{\sqrt{H(X_1)H(Y)}}
<
\frac{I(X_2;Y)}
{\sqrt{H(X_2)H(Y)}}.
\]

Thus, symmetric NMI can favor lower-entropy variables and cannot be interpreted unambiguously as information recovery \cite{jerdee2025}.

\subsection{Asymmetric Normalization}

Jerdee et al.\ propose an asymmetric normalization \cite{jerdee2025}:

\[
\mathrm{NMI}_{\mathrm{asym}}(X;Y)
=
\frac{I(X;Y)}{H(Y)}.
\]

This preserves ranking:

\[
I(X_1;Y) > I(X_2;Y)
\;\Rightarrow\;
\frac{I(X_1;Y)}{H(Y)}
>
\frac{I(X_2;Y)}{H(Y)}.
\]

Asymmetric normalization is appropriate when there is a clear reference or ground-truth variable and when interpretability and comparability are the primary goals \cite{jerdee2025}. Symmetric and asymmetric normalizations answer fundamentally different questions: symmetric NMI measures mutual similarity, whereas asymmetric NMI quantifies how much of a reference variable is explained or recovered by another. The present work adopts asymmetric normalization primarily because of its ranking-preserving property rather than any particular boundedness guarantee. 

\subsection{KSG-Based NMI Estimation}

Classical estimators rely on explicit density estimation and suffer in high-dimensional or nonlinear settings. For example, $k$NN-based MI estimators work well in low dimensions but degrade rapidly as dimensionality increases \cite{kraskov2004estimating}.

A common approach for continuous NMI estimation combines KSG mutual information estimation with nearest-neighbor entropy estimation \cite{kraskov2004estimating,nagel2024accurate}. Although effective in low-dimensional settings, these estimators may suffer from increasing variance and numerical instability as dimensionality grows \cite{tuononen2025stability}.

\subsection{Mutual Information Neural Estimation}

MINE is based on the Donsker--Varadhan (DV) variational representation of the Kullback--Leibler divergence \cite{donsker1983asie,belghazi2018mine}. The DV representation reformulates KL divergence as an optimization problem over a family of functions. This is particularly useful in neural estimation because the optimizing function can be approximated by a neural network, allowing divergence and mutual information to be estimated directly from samples \cite{belghazi2018mine,mcallester2018limitations}.

\[
D_{\mathrm{KL}}(p \| m)
=
\sup_{T \in \mathcal{F}}
\left(
\mathbb{E}_{p}[T(X)]
-
\log
\mathbb{E}_{m}
\left[
e^{T(X)}
\right]
\right).
\]
\subsection{Neural Entropy Estimation}

The proposed estimator recovers marginal and joint entropies from neural divergence estimates relative to uniform reference distributions. This approach is inspired by Neural Entropic Estimation (NEE) \cite{chan2019nee} and the relative-entropy formulation of continuous entropy \cite{jaynes1968prior}. Related divergence-based approaches have also been investigated for neural density and entropy estimation \cite{park2021ddde}. Following NEE, entropy is estimated indirectly by measuring the divergence between the target distribution and a reference distribution \cite{chan2019nee}. In this work, the reference distribution is chosen to be uniform over the observed support of the data. This choice is convenient because the reference density and its associated volume term are available in closed form, allowing entropy to be recovered directly from divergence estimates \cite{jaynes1968prior,chan2019nee}.

Let $m_X$ denote a uniform reference distribution defined over the observed support of $X$. The reference distribution acts as the reference measure with respect to which entropy is recovered through KL divergence. The corresponding support volume is

\[
V_X
=
\prod_i
(x_i^{\max}-x_i^{\min}).
\]

Since the reference density is uniform,

\[
m_X(x)
=
\frac{1}{V_X}.
\]

The divergence between the empirical distribution and the reference distribution is

\[
D_X
=
D_{\mathrm{KL}}
\left(
p_X
\;\|\;
m_X
\right).
\]

Expanding the KL divergence gives

\[
D_X
=
\int p_X(x)
\log
\frac{p_X(x)}
     {m_X(x)}
\,dx.
\]

Substituting the uniform reference density yields

\[
D_X
=
\int p_X(x)\log p_X(x)\,dx
+
\log V_X.
\]

Since differential entropy is defined as \cite{cover2006elements}

\[
H(X)
=
-
\int p_X(x)\log p_X(x)\,dx,
\]

entropy can be recovered as

\[
H(X)
=
\log V_X
-
D_X.
\]

Using neural divergence estimates results in the entropy estimator

\[
\widehat H(X)
=
\log V_X
-
\widehat D_X.
\]

Similarly,

\[
\widehat H(Y)
=
\log V_Y
-
\widehat D_Y,
\]

and

\[
\widehat H(X,Y)
=
\log V_{XY}
-
\widehat D_{XY}.
\]

Thus, entropy estimation does not require explicit density estimation. Instead, entropy is recovered directly from neural divergence estimates and the support volumes of the corresponding uniform reference distributions \cite{chan2019nee,jaynes1968prior}. The support volumes are estimated from the observed sample ranges. For a $d$-dimensional variable, the reference volume is computed as the product of side lengths of the smallest axis-aligned box containing the observed samples. In practice, the support volume is estimated from the observed sample range rather than the true support of the underlying distribution. The use of observed sample ranges provides a practical finite-sample approximation of the reference support, but may introduce sensitivity to sample size and extreme observations.

\section{Scale-Invariant Estimation}

Meaningful information-theoretic measures should be invariant under changes of coordinates and measurement units. Without this property, estimators can produce distorted values that depend on the units of measurement or the representation of the variables. Scale-invariant normalization and estimation are therefore necessary for meaningful MI and NMI estimation \cite{nagel2024accurate}.

\subsection{Entropies}

For a continuous random variable $X$, differential entropy is defined as \cite{cover2006elements}

\[
H_d(X)
=
-\int p_X(x)\log p_X(x)\,dx.
\]

Under scaling $x' = ax$ with $a>0$, the Jacobian determinant is

\[
\left|\frac{\partial x}{\partial x'}\right|
=
\frac{1}{a},
\]

and differential entropy transforms as

\[
H_d(X')
=
H_d(X)
+
\log a.
\]

Thus, differential entropy is not invariant under scaling \cite{cover2006elements,jaynes1968prior}.

Jaynes introduced the continuous analogue of discrete entropy as a relative entropy with respect to a reference measure $m(x)$ \cite{jaynes1968prior,nagel2024accurate}:

\[
H_r(X)
=
-
\int
p_X(x)
\log
\frac{p_X(x)}
     {m_X(x)}
\,dx.
\]

Under scaling, both $p_X$ and $m_X$ transform identically, making $H_r$ invariant \cite{jaynes1968prior,nagel2024accurate}. For multivariate systems, the invariant reference measure can be chosen to factorize as

\[
m_{XY}(x,y)
=
m_X(x)m_Y(y),
\]

which preserves invariance under coordinate transformations \cite{nagel2024accurate}.

\subsection{Invariance of Mutual Information}

Although differential entropies are not invariant, mutual information remains invariant because it is fundamentally a Kullback--Leibler divergence \cite{cover2006elements}:

\[ 
\begin{aligned} 
I_d(X;Y) 
&= 
H_d(X)+H_d(Y) - H_d(X,Y) \\ 
&= 
H_r(X)+H_r(Y) - H_r(X,Y). 
\end{aligned} 
\]

Thus, mutual information remains invariant even though differential entropy does not.

\subsection{Invariant Normalized Mutual Information}

Normalization based on differential entropies is not invariant:

\[ \mathrm{NMI}^{\mathrm{sym}}_d(X,Y) = \frac{I(X;Y)} {\sqrt{H_d(X)H_d(Y)}}, \]

\[ \mathrm{NMI}^{\mathrm{asym}}_d(X;Y) = \frac{I(X;Y)} {H_d(Y)}. \]

Under scaling, the denominators change, breaking invariance \cite{nagel2024accurate}.

By contrast, normalization based on relative entropies is invariant:

\[
\mathrm{NMI}^{\mathrm{sym}}_r(X,Y)
=
\frac{I(X;Y)}
{\sqrt{H_r(X)H_r(Y)}},
\]

\[
\mathrm{NMI}^{\mathrm{asym}}_r(X;Y)
=
\frac{I(X;Y)}
{H_r(Y)}.
\]

These expressions provide the theoretical motivation for invariant normalized mutual information estimation and serve as the basis for the proposed framework \cite{nagel2024accurate}.

\section{Proposed Method}

We propose Neural Mutual Information and Entropy Estimation (NMINE), a fully neural estimator for normalized mutual information between continuous multidimensional variables.

The proposed framework combines neural divergence estimation with entropy recovery from uniform reference distributions. Unlike nearest-neighbor-based approaches, all quantities required for mutual information and normalized mutual information estimation are obtained through neural variational estimation.

\subsection{Neural Divergence Estimation}

Let $m_X$, $m_Y$, and $m_Xm_Y$ denote uniform reference distributions defined over the supports of $X$, $Y$, and $(X,Y)$, respectively.

Following the Donsker--Varadhan representation \cite{donsker1983asie} used in MINE \cite{belghazi2018mine}, each Kullback--Leibler divergence is estimated using a neural critic:

\[
D_{\mathrm{KL}}(p\|m)
=
\sup_{T\in\mathcal{F}}
\left(
\mathbb{E}_p[T]
-
\log
\mathbb{E}_m[e^T]
\right).
\]

Three neural critics are trained independently:

\[
D_{XY}
=
D_{\mathrm{KL}}
\left(
p_{XY}
\;\|\;
m_Xm_Y
\right),
\]

\[
D_X
=
D_{\mathrm{KL}}
\left(
p_X
\;\|\;
m_X
\right),
\]

and

\[
D_Y
=
D_{\mathrm{KL}}
\left(
p_Y
\;\|\;
m_Y
\right).
\]

The corresponding neural estimates are denoted by

\[
\widehat D_{XY},
\qquad
\widehat D_X,
\qquad
\widehat D_Y.
\]

\subsection{Mutual Information Estimation}

Using the relative-entropy formulation of mutual information, the NMINE mutual information estimate is obtained as

\[
\widehat I(X;Y)
=
\widehat D_{XY}
-
\widehat D_X
-
\widehat D_Y .
\]

This formulation follows directly from the decomposition

\[
I(X;Y)
=
D_{\mathrm{KL}}(p_{XY}\|m_Xm_Y)
-
D_{\mathrm{KL}}(p_X\|m_X)
-
D_{\mathrm{KL}}(p_Y\|m_Y).
\]

This decomposition follows directly from the relative-entropy formulation and does not require explicit density estimation.
Because the three divergences are estimated independently using variational objectives, the resulting mutual-information estimate is not itself a variational bound. Analyzing the bias and variance induced by this decomposition remains an important direction for future work.

\subsection{Neural Entropy Estimation}

Marginal and joint entropies are recovered from the estimated divergences.

Let

\[
V_X
=
\prod_i
(x_i^{\max}-x_i^{\min}),
\]

\[
V_Y
=
\prod_i
(y_i^{\max}-y_i^{\min}),
\]

and

\[
V_{XY}
=
V_XV_Y
\]

denote the volumes of the uniform reference supports.

The entropy estimates are obtained as

\[
\widehat H(X)
=
\log V_X
-
\widehat D_X,
\]

\[
\widehat H(Y)
=
\log V_Y
-
\widehat D_Y,
\]

and

\[
\widehat H(X,Y)
=
\log V_{XY}
-
\widehat D_{XY}.
\]

This approach is inspired by Neural Entropic Estimation (NEE) \cite{chan2019nee} and the relative-entropy formulation of continuous entropy \cite{jaynes1968prior}. Related divergence-based approaches have also employed the Donsker--Varadhan representation together with uniform reference distributions for density and entropy-related estimation tasks \cite{park2021ddde}. Thus, entropy estimation does not require explicit density estimation. Instead, entropy is recovered directly from neural divergence estimates and the support volumes of the corresponding uniform reference distributions. The theoretical formulation follows the relative-entropy framework of Jaynes \cite{jaynes1968prior}, where entropy is defined relative to a reference measure. In practice, however, the reference support is approximated from finite samples using observed data ranges. Consequently, the proposed implementation should be viewed as a finite-sample approximation of the ideal relative-entropy formulation and may exhibit sensitivity to sample size and extreme observations.

\subsection{Normalized Mutual Information Estimation}

Following the asymmetric normalization proposed by Jerdee et al. \cite{jerdee2025}, the final NMINE estimate is computed as \[ \widehat{\mathrm{NMI}}(X;Y) = \frac{\widehat I(X;Y)} {\widehat H(Y)}. \] This normalization preserves ranking with respect to mutual information and admits an interpretation in terms of information recovery. Following Jerdee et al. \cite{jerdee2025}, we adopt asymmetric normalization because it preserves the ordering induced by mutual information. Unlike symmetric normalization schemes, the asymmetric formulation admits a direct interpretation in terms of information recovery, quantifying the fraction of information contained in a reference variable that is explained by another variable.

The proposed estimator uses neural divergence estimation for both the mutual information and entropy terms, yielding a fully neural pipeline for normalized mutual information estimation.

\subsection{Implementation}

The neural critics were implemented as fully connected feedforward networks consisting of two hidden layers with 128 units and ReLU activations, followed by a linear output layer (Table~\ref{tab:critic_architecture}). Separate networks were trained for the joint distribution and the two marginal distributions.

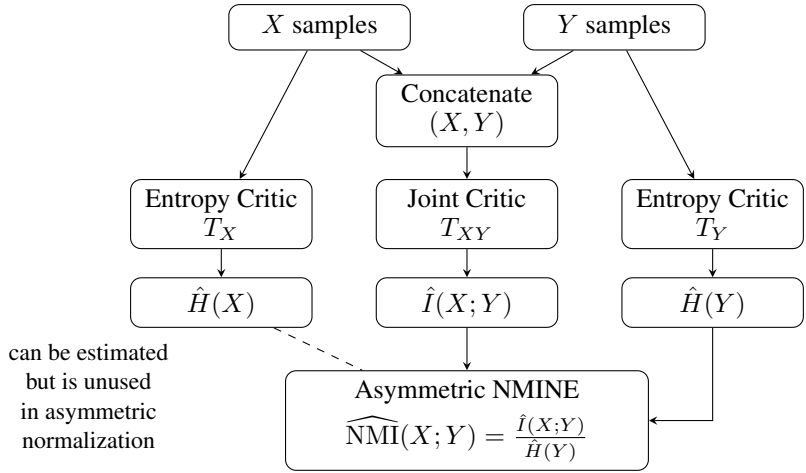
\begin{figure*}[t] 
\centering 
\begin{tikzpicture}[ scale=0.65, every node/.style={align=center}, block/.style={ draw, rounded corners, minimum width=2.4cm, minimum height=0.6cm }, >=stealth ] 
\node[block] (x) at (-3,0.5) {$X$ samples}; 
\node[block] (y) at (3,0.5) {$Y$ samples}; 
\node[block] (concat) at (0,-1.2) 
{Concatenate\\$(X,Y)$}; 
\draw[->] (x) -- (concat); 
\draw[->] (y) -- (concat); 
\node[block] (joint) at (0,-3.3) 
{Joint Critic\\$T_{XY}$}; 
\node[block] (mi) at (0,-5.1) 
{$\hat I(X;Y)$}; 
\draw[->] (concat) -- (joint); 
\draw[->] (joint) -- (mi);
\node[block] (tx) at (-5,-3.3) 
{Entropy Critic\\$T_X$}; 
\node[block] (hx) at (-5,-5.1) 
{$\hat H(X)$}; 
\draw[->] (x) -- (tx); 
\draw[->] (tx) -- (hx);
\node[block] (ty) at (5,-3.3) 
{Entropy Critic\\$T_Y$}; 
\node[block] (hy) at (5,-5.1) 
{$\hat H(Y)$}; 
\draw[->] (y) -- (ty); 
\draw[->] (ty) -- (hy); 
\node[block, minimum width=4.8cm] (nmi) at (0,-7.5) { Asymmetric NMINE\\[1mm] $\widehat{\mathrm{NMI}}(X;Y) = \frac{\hat I(X;Y)} {\hat H(Y)}$ }; 
\draw[->] (mi) -- (nmi); 
\draw[->] (hy) |- (nmi);
\draw[dashed] (hx) -- (nmi); 
\node at (-7.7,-7.0) 
{
    \footnotesize can be estimated \\
    \footnotesize but is unused \\
    \footnotesize in asymmetric \\
    \footnotesize normalization
}; 
\end{tikzpicture} 
\caption{ Architecture of the proposed NMINE framework. Input samples $X$ and $Y$ are used both for marginal entropy estimation and for constructing joint samples $(X,Y)$, which are provided to the mutual-information critic. Although both marginal entropies are estimated, the final estimator uses asymmetric normalization and computes $\widehat{\mathrm{NMI}}(X;Y)=\hat I(X;Y)/\hat H(Y)$. } \label{fig:nmine_architecture} 
\end{figure*}

\begin{table}[t]
    \centering
    \caption{Neural critic architecture used in all experiments.}
    \label{tab:critic_architecture}
    \begin{tabular}{ll}
        \hline
        Layer & Configuration \\
        \hline
        Input          & Variable dimensionality \\
        Hidden layer 1 & Fully connected (128 units) + ReLU \\
        Hidden layer 2 & Fully connected (128 units) + ReLU \\
        Output layer   & Fully connected (1 unit), linear activation \\
        \hline
    \end{tabular}
\end{table}

Optimization was performed using the Adam optimizer \cite{kingma2015adam} with a learning rate of $10^{-5}$. The mutual-information critic was trained for 500 epochs using a batch size of 256. The entropy critics were trained independently for 100 epochs using the same batch size. No additional variance-reduction techniques such as exponential moving averages or critic clipping were used.

During training, empirical samples were contrasted against uniformly sampled reference distributions defined over the observed support of each variable. The resulting divergence estimates were subsequently combined to obtain estimates of mutual information, entropy, and normalized mutual information.

\section{Experimental setup}

\subsection{ Synthetic Gaussian Data}
 Data were generated from a zero-mean multivariate Gaussian distribution with covariance matrix \[ \Sigma= \begin{bmatrix} I & \rho I\\ \rho I & I \end{bmatrix}, \] where $\rho$ controls the correlation between the variables. Experiments were performed for dimensions $d\in\{1,2,4,8\}$ and correlation coefficients $\rho\in\{0,0.1,\ldots,0.9,0.95\}$. For each parameter configuration, 5000 samples were generated and the experiment was repeated using 10 random seeds. 

Ground-truth mutual information and normalized mutual information values were computed analytically from the corresponding Gaussian model using the same asymmetric normalization employed by NMINE. 

We compare NMINE against a KSG-based normalized mutual-information estimator composed of Kraskov mutual information estimation and nearest-neighbor entropy estimation. The KSG estimator used $k=5$ nearest neighbors. Performance was evaluated using mean absolute error (MAE) between estimated and analytical normalized mutual information values.

\section{Experimental Results} 
\begin{figure}[t] 
\centering 
\includegraphics[width=\linewidth]{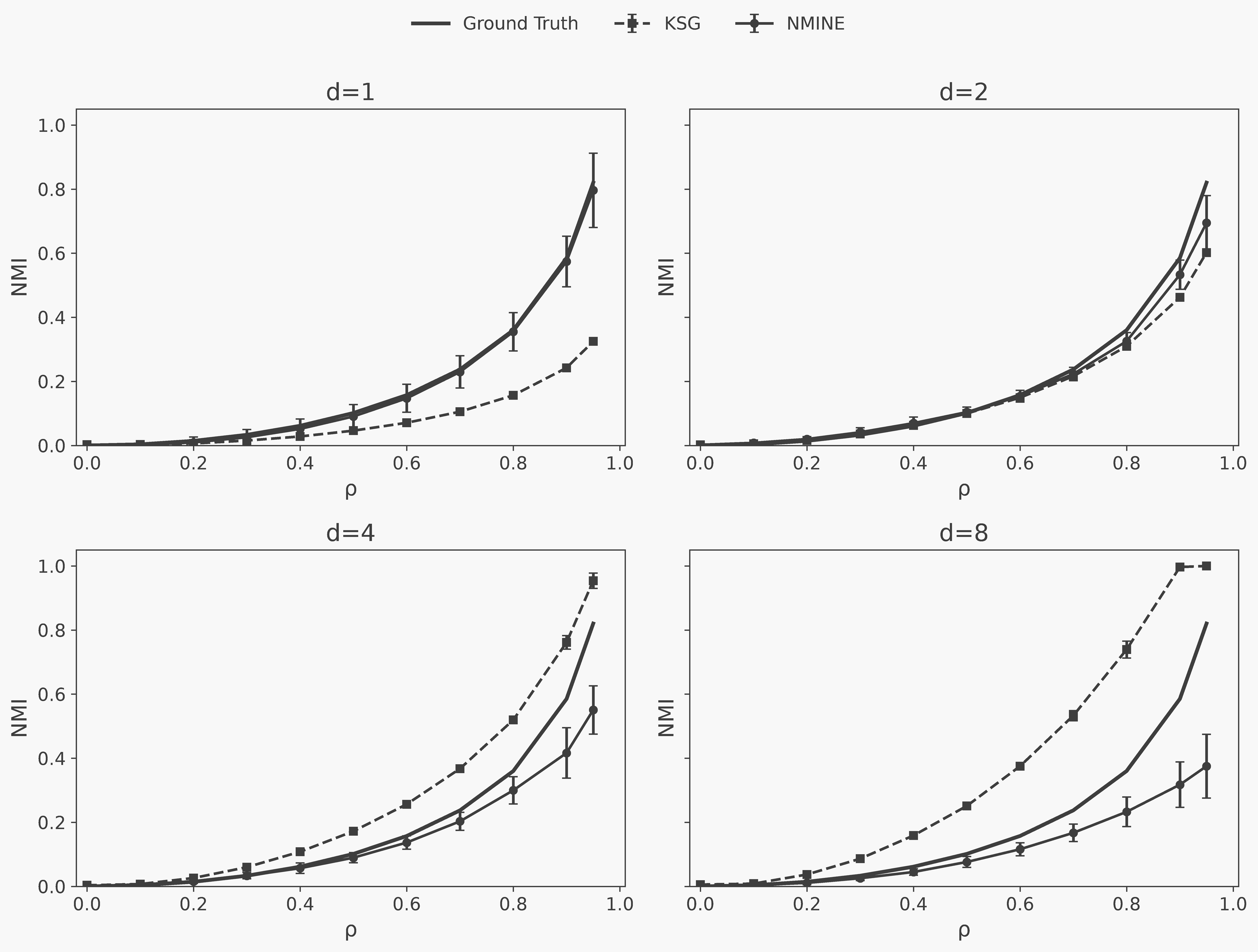} 
\caption{ Estimated normalized mutual information (NMI) as a function of correlation coefficient $\rho$ for dimensions $d \in \{1,2,4,8\}$. Error bars represent standard deviations across random seeds. } 
\label{fig:nmi_results} 
\end{figure}

Figure~\ref{fig:nmi_results} compares NMINE against the KSG-based baseline across dimensions one to eight. The proposed estimator achieves lower overall estimation error than the baseline estimator. The performance gap becomes increasingly visible in higher-dimensional settings, where nearest-neighbor methods are known to experience increased estimation variance and numerical instability. These results support the hypothesis that normalized mutual information can be estimated effectively using a fully neural pipeline. 

As shown in Figure~\ref{fig:nmi_results}, NMINE closely follows the theoretical NMI curve in low-dimensional settings ($d=1$ and $d=2$). As dimensionality increases, both estimators become less accurate; however, their failure modes differ substantially. KSG increasingly overestimates the theoretical NMI, whereas NMINE exhibits a systematic underestimation bias. The most pronounced differences are observed in the highest-dimensional setting ($d=8$), where KSG approaches the upper limit of the normalized scale for strong correlations while NMINE remains conservative. Although NMINE underestimates the true NMI in these cases, it preserves the correct monotonic relationship between correlation strength and normalized mutual information.

\begin{table}[t] \centering \footnotesize \setlength{\tabcolsep}{4pt} \caption{ Mean absolute error (MAE) relative to theoretical NMI values. Results are reported as mean $\pm$ standard deviation across random seeds. The final column reports the relative MAE reduction achieved by NMINE compared to KSG. } \label{tab:mae} \begin{tabular}{lccc} \hline Dimension & KSG MAE & NMINE MAE & Reduction (\%) \\ \hline 1 & 0.1387 $\pm$ 0.0127 & 0.0346 $\pm$ 0.0246 & 73.9 $\pm$ 20.5 \\ 2 & 0.0410 $\pm$ 0.0023 & 0.0280 $\pm$ 0.0133 & 32.7 $\pm$ 29.7 \\ 4 & 0.0830 $\pm$ 0.0090 & 0.0568 $\pm$ 0.0156 & 30.7 $\pm$ 20.6 \\ 8 & 0.1746 $\pm$ 0.0070 & 0.0920 $\pm$ 0.0242 & 47.3 $\pm$ 13.9 \\ \hline \end{tabular} \end{table}

To quantify estimation accuracy, Table~\ref{tab:mae} reports the mean absolute error (MAE) relative to the theoretical NMI values. NMINE achieved lower MAE than the KSG baseline across all evaluated dimensionalities. Relative error reductions ranged from 30.7\% to 74.0\%. A paired t-test comparing the absolute estimation errors of NMINE and KSG across all evaluation runs revealed a statistically significant difference in favor of NMINE ($t=9.31$, $p=7.89\times10^{-19}$). This result provides strong evidence that NMINE achieves lower overall estimation error than the KSG-based baseline.

The systematic underestimation observed for NMINE in higher-dimensional settings is consistent with previously reported challenges in neural mutual-information estimation. In particular, strong statistical dependencies are often more difficult to estimate accurately than weak dependencies \cite{belghazi2018mine}. Moreover, theoretical analyses have demonstrated fundamental limitations of variational mutual-information estimators when mutual information is large relative to the available sample size \cite{mcallester2018}. Moreover, theoretical analyses have demonstrated fundamental limitations of variational mutual-information estimators when mutual information is large relative to the available sample size \cite{mcallester2018}.

\subsection{Preliminary Student-\textit{t} Experiments} To assess robustness beyond Gaussian data, we conducted preliminary experiments using multivariate Student-$t$ distributions. Although analytical ground-truth NMI values were not available, NMINE exhibited a generally monotonic response to increasing correlation across evaluated dimensions $d\in\{1,2,4,8\}$. In addition, NMINE showed lower variability than the KSG baseline in several moderate-correlation settings. These results suggest that the proposed framework may remain applicable in heavy-tailed settings, although a more comprehensive evaluation is left for future work.

\section{Discussion and Limitations} 

The experimental results support the central hypothesis of this work: normalized mutual information can be estimated effectively using a fully neural framework. The proposed estimator combines complementary strengths of MINE-based mutual information estimation and MI-NEE-inspired entropy estimation. 

Potential applications include interpretable machine learning, where normalized mutual information may be used to compare information content across neural representations \cite{tuononen2025receiver}, and molecular dynamics, where multidimensional dependency analysis is required \cite{nagel2024accurate}. The method may also be relevant to representation learning and out-of-distribution detection \cite{hendrycks2017baseline}. Distributional shifts may alter dependency structures between latent variables and semantic factors, suggesting normalized mutual information as a potential dependency signal for detecting such shifts \cite{chen2018isolating,ramakrishna2021ood}. These applications remain speculative and are left for future work. 

The current implementation uses the standard Donsker--Varadhan objective without additional variance reduction techniques such as SMILE, exponential moving averages, or alternative contrastive objectives. Investigating such techniques remains an important direction for future work. First, the mutual information and entropy estimators are trained separately. Future work should investigate joint optimization of both components. 

Second, entropy estimation requires selecting the support of the uniform reference distribution. The proposed formulation assumes a reference distribution defined on the observed sample support rather than on the true support of the underlying distribution. For distributions with unbounded support, such as Gaussians, this introduces a finite-sample approximation whose effect should be investigated further. The influence of sample-dependent support estimation on entropy recovery and normalization also warrants further study. 

Third, Gaussian data provide only a controlled benchmark. Preliminary experiments on Student-$t$ data further suggest that the proposed framework may remain applicable outside Gaussian settings, although a systematic evaluation of non-Gaussian distributions is left for future work. Future evaluation should also include nonlinear and real-world datasets. 

Finally, the proposed method incurs additional computational cost. Unlike nearest-neighbor estimators, NMINE requires training multiple neural networks, increasing computational requirements and training time.

\section{Conclusions}

We proposed a fully neural estimator for normalized mutual information between continuous multidimensional variables. The method combines a MINE-based mutual information estimator with MI-NEE-inspired marginal entropy estimators. Mutual information is estimated using a neural Donsker–Varadhan objective, while marginal entropies are estimated by comparing empirical marginals to uniform reference distributions and recovering entropy from the resulting divergence estimates.

Experiments on Gaussian data from one to eight dimensions show that the proposed estimator is more accurate than a KSG-based normalized mutual information baseline. These findings demonstrate that normalized mutual information can be estimated effectively within a fully neural framework, providing a viable alternative to traditional nearest-neighbor-based estimators. Preliminary experiments on heavy-tailed Student-$t$ distributions exhibited qualitatively similar monotonic behavior, suggesting that the proposed framework may remain applicable beyond Gaussian settings.

Future work will focus on joint training of the MI and entropy estimators, evaluation on nonlinear and non-Gaussian dependencies, and applications to interpretable machine learning, molecular dynamics, disentangled representation learning, and out-of-distribution detection.

\bibliography{references}

@book{cover2006elements,
  title={Elements of Information Theory},
  author={Cover, Thomas M. and Thomas, Joy A.},
  year={2006},
  publisher={Wiley}
}

@article{kraskov2004estimating,
  title={Estimating Mutual Information},
  author={Kraskov, Alexander and Stogbauer, Harald and Grassberger, Peter},
  journal={Physical Review E},
  volume={69},
  number={6},
  pages={066138},
  year={2004}
}

@inproceedings{belghazi2018mine,
  title={Mutual Information Neural Estimation},
  author={Belghazi, Mohamed Ishmael and Baratin, Aristide and Rajeshwar, Sai and Ozair, Sherjil and Bengio, Yoshua and Courville, Aaron and Hjelm, Devon},
  booktitle={International Conference on Machine Learning},
  pages={531--540},
  year={2018}
}

@article{chan2019nee,
  title={Neural Entropic Estimation},
  author={Chan, Jeffrey and Al-Bashabsheh, Ali and Ebrahimi, Jamal and Koyluoglu, Ozan and Varshney, Pramod},
  journal={Entropy},
  volume={21},
  number={10},
  pages={1006},
  year={2019}
}

@article{paninski2003estimation,
  title={Estimation of Entropy and Mutual Information},
  author={Paninski, Liam},
  journal={Neural Computation},
  volume={15},
  number={6},
  pages={1191--1253},
  year={2003}
}

@article{gao2015efficient,
  title={Efficient Estimation of Mutual Information for Strongly Dependent Variables},
  author={Gao, Shuyang and Ver Steeg, Greg and Galstyan, Aram},
  journal={AISTATS},
  year={2015}
}

@inproceedings{oord2018representation,
  title={Representation Learning with Contrastive Predictive Coding},
  author={van den Oord, Aaron and Li, Yazhe and Vinyals, Oriol},
  booktitle={arXiv preprint arXiv:1807.03748},
  year={2018}
}

@article{hjelm2019learning,
  title={Learning Deep Representations by Mutual Information Estimation and Maximization},
  author={Hjelm, Devon and Fedorov, Alex and Lavoie-Marchildon, Samuel and Grewal, Karan and Bachman, Philip and Trischler, Adam and Bengio, Yoshua},
  journal={ICLR},
  year={2019}
}

@article{vinh2010information,
  title={Information Theoretic Measures for Clusterings Comparison},
  author={Vinh, Nguyen Xuan and Epps, Julien and Bailey, James},
  journal={Journal of Machine Learning Research},
  volume={11},
  pages={2837--2854},
  year={2010}
}

@article{jaynes1968prior,
  title={Prior Probabilities},
  author={Jaynes, Edwin T.},
  journal={IEEE Transactions on Systems Science and Cybernetics},
  volume={4},
  number={3},
  pages={227--241},
  year={1968}
}

@article{nagel2024accurate,
  author = {Nagel, Daniel and Diez, Georg and Stock, Gerhard},
  title = {Accurate Estimation of the Normalized Mutual Information of Multidimensional Data},
  journal = {The Journal of Chemical Physics},
  volume = {161},
  number = {5},
  pages = {054108},
  year = {2024},
  doi = {10.1063/5.0217960}
}

@article{tuononen2025stability,
  author = {Tuononen, Marko and Hautam{\"a}ki, Ville},
  title = {Improving Numerical Stability of Normalized Mutual Information Estimator on High Dimensions},
  journal = {IEEE Signal Processing Letters},
  volume = {32},
  pages = {2783--2787},
  year = {2025},
  doi = {10.1109/LSP.2025.3588026}
}

@article{jerdee2025,
  author = {Jerdee, Maximilian and Kirkley, Alec and Newman, M. E. J.},
  title = {Normalized Mutual Information Is a Biased Measure for Classification and Community Detection},
  journal = {Nature Communications},
  volume = {16},
  pages = {11268},
  year = {2025}
}

@inproceedings{tuononen2025receiver,
  author = {Tuononen, Marko and Korpi, Dani and Hautam{\"a}ki, Ville},
  title = {Interpreting Deep Neural Network-Based Receiver Under Varying Signal-To-Noise Ratios},
  booktitle = {ICASSP 2025 -- IEEE International Conference on Acoustics, Speech and Signal Processing},
  pages = {1--5},
  year = {2025},
  doi = {10.1109/ICASSP49660.2025.10888682}
}

@article{mcallester2018,
  title={Formal Limitations on the Measurement of Mutual Information},
  author={McAllester, David and Stratos, Karl},
  journal={arXiv preprint arXiv:1811.04251},
  year={2018}
}

@inproceedings{chen2018isolating,
  title={Isolating Sources of Disentanglement in Variational Autoencoders},
  author={Chen, Ricky T. Q. and Li, Xuechen and Grosse, Roger and Duvenaud, David},
  booktitle={Advances in Neural Information Processing Systems},
  volume={31},
  year={2018}
}

@article{ramakrishna2021ood,
  title={Out-of-Distribution Detection in Deep Learning: A Survey},
  author={Ramakrishna, Sai V. S. and Uhler, Caroline and others},
  journal={arXiv preprint arXiv:2110.11334},
  year={2021}
}

@inproceedings{sanchez2020disentangled,
  title={Learning Disentangled Representations via Mutual Information Estimation},
  author={Sanchez, Eduardo Hugo and Serrurier, Mathieu and Ortner, Mathias},
  booktitle={European Conference on Computer Vision Workshops},
  year={2020}
}

@inproceedings{chen2016infogan,
  title={InfoGAN: Interpretable Representation Learning by Information Maximizing Generative Adversarial Nets},
  author={Chen, Xi and Duan, Yan and Houthooft, Rein and Schulman, John and Sutskever, Ilya and Abbeel, Pieter},
  booktitle={Advances in Neural Information Processing Systems},
  volume={29},
  year={2016}
}

@inproceedings{strehl2002cluster,
  title={Cluster Ensembles: A Knowledge Reuse Framework for Combining Multiple Partitions},
  author={Strehl, Alexander and Ghosh, Joydeep},
  booktitle={AAAI Conference on Artificial Intelligence},
  pages={93--98},
  year={2002}
}

@inproceedings{hendrycks2017baseline,
  title={A Baseline for Detecting Misclassified and Out-of-Distribution Examples in Neural Networks},
  author={Hendrycks, Dan and Gimpel, Kevin},
  booktitle={International Conference on Learning Representations},
  year={2017}
}

@article{kingma2015adam,
  title={Adam: A Method for Stochastic Optimization},
  author={Kingma, Diederik P. and Ba, Jimmy},
  journal={International Conference on Learning Representations},
  year={2015}
}

@article{donsker1983asie,
  title={Asymptotic Evaluation of Certain Markov Process Expectations for Large Time},
  author={Donsker, Monroe D. and Varadhan, S. R. S.},
  journal={Communications on Pure and Applied Mathematics},
  volume={36},
  number={2},
  pages={183--212},
  year={1983}
}

@article{park2021ddde,
  title={Deep Data Density Estimation through Donsker--Varadhan Representation},
  author={Park, Seonho and Pardalos, Panos M.},
  journal={arXiv preprint arXiv:2104.06612},
  year={2021}
}

@article{shannon1948mathematical,
  author = {Shannon, Claude E.},
  title = {A Mathematical Theory of Communication},
  journal = {The Bell System Technical Journal},
  volume = {27},
  number = {3},
  pages = {379--423},
  year = {1948},
  month = {July},
  doi = {10.1002/j.1538-7305.1948.tb01338.x}
}

@article{mcallester2018limitations,
  title={Formal Limitations on the Measurement of Mutual Information},
  author={McAllester, David and Stratos, Karl},
  journal={arXiv preprint arXiv:1811.04251},
  year={2018}
}

\end{document}